\documentclass{article}
\usepackage{microtype}
\usepackage{graphicx}
\usepackage{booktabs} % for professional tables
\usepackage{hyperref}
\usepackage{float}
\usepackage{subfig}
\usepackage[utf8]{inputenc}
\usepackage{authblk}
\usepackage{amsmath,amssymb,bm}
\usepackage[normalem]{ulem}
\usepackage[dvipsnames]{xcolor}
\usepackage{graphicx}
\usepackage{bbm}
\usepackage{mathtools}
\usepackage{enumitem}
\usepackage{mathtools}

\usepackage{multirow}
\usepackage{url}
\PassOptionsToPackage{hyphens}{url}
\usepackage{hyperref}

\DeclarePairedDelimiter\norm{\lVert}{\rVert}

\newcommand{\quotes}[1]{``#1"}

\renewcommand{\vec}[1]{\bm{#1}}

\newcommand{\oneNorm}[2][1]{%
    \ensuremath{{}_{\phantom{#1}}\norm{\vec{#2}}_{#1}}%
}
\newcommand{\twoNorm}[2][2]{%
    \ensuremath{{}_{\phantom{#1}}\norm{\vec{#2}}_{#1}}%
}

\usepackage[accepted]{mlsys2022}

\mlsystitlerunning{Quantized Neural Networks for Low-Precision Accumulation with Guaranteed Overflow Avoidance}

\begin{document}

\twocolumn[
\mlsystitle{Quantized Neural Networks for Low-Precision Accumulation with Guaranteed Overflow Avoidance}

\begin{mlsysauthorlist}
\mlsysauthor{Ian Colbert}{amd_san_diego}
\mlsysauthor{Alessandro Pappalardo}{amd_dublin}
\mlsysauthor{Jakoba Petri-Koenig}{amd_dublin}
\end{mlsysauthorlist}

\mlsysaffiliation{amd_san_diego}{AMD SW Technology Team, San Diego, California, USA}
\mlsysaffiliation{amd_dublin}{AMD AECG Research Labs, Dublin, Ireland}

\mlsyscorrespondingauthor{Ian Colbert}{ian.colbert@amd.com}

\mlsyskeywords{Machine Learning, Quantization, Deep Neural Networks, FPGA Design, Programmable Hardware, Inference Optimization}

\vskip 0.3in

\begin{abstract}
Quantizing the weights and activations of neural networks significantly reduces their inference costs, often in exchange for minor reductions in model accuracy.
This is in large part due to compute and memory cost savings in operations like convolutions and matrix multiplications, whose resulting products are typically accumulated into high-precision registers, referred to as accumulators.
While many researchers and practitioners have taken to leveraging low-precision representations for the weights and activations of a model, few have focused attention on reducing the size of accumulators.
Part of the issue is that accumulating into low-precision registers introduces a high risk of numerical overflow which, due to wraparound arithmetic, can significantly degrade model accuracy.
In this work, we introduce a quantization-aware training algorithm that guarantees avoiding numerical overflow when reducing the precision of accumulators during inference.
We leverage weight normalization as a means of constraining parameters during training using accumulator bit width bounds that we derive.
We evaluate our algorithm across multiple quantized models that we train for different tasks, showing that our approach can reduce the precision of accumulators while maintaining model accuracy with respect to a floating-point baseline.
We then show that this reduction translates to increased design efficiency for custom FPGA-based accelerators.
Finally, we show that our algorithm not only constrains weights to fit into an accumulator of user-defined bit width, but also increases the sparsity and compressibility of the resulting weights.
Across all of our benchmark models trained with 8-bit weights and activations, we observe that constraining the hidden layers of quantized neural networks to fit into 16-bit accumulators yields an average 98.2\% sparsity with an estimated compression rate of 46.5x all while maintaining 99.2\% of the floating-point performance.
\end{abstract}
]

\printAffiliationsAndNotice{}

\section{Introduction}
\label{sec:introduction}

Quantization is the process of reducing the range and precision of the numerical representation of data.
Among the many techniques used to reduce the inference costs of neural networks (NNs), integer quantization is one of the most widely applied in practice~\cite{gholami2021survey}.
The reduction in compute and memory requirements resulting from low-precision quantization provides increased throughput, power savings, and resource efficiency, usually in exchange for minor reductions in model accuracy~\cite{hubara2017quantized}.
During inference, information is propagated through the layers of an NN, where most of the compute workload is concentrated in the multiply-and-accumulates (MACs) of operators such as convolutions and matrix multiplications.
It has been shown that reducing the bit width of the accumulator can increase throughput and bandwidth efficiency for general-purpose processors by creating more opportunities to increase parallelism~\cite{de2020quantization,ni2021wrapnet,xie2021overflow}.
However, exploiting such an optimization is non-trivial, as doing so incurs a high risk of overflow which can introduce numerical errors that significantly degrade model accuracy due to wraparound twos-complement arithmetic~\cite{ni2021wrapnet}.

Previous work has sought to either reduce the risk of numerical overflow~\cite{xie2021overflow, sakr2019accumulation} or mitigate its impact on model accuracy~\cite{ni2021wrapnet}.
In this work, we train quantized NNs (QNNs) to avoid numerical overflow altogether when using low-precision accumulators during inference.
To fully exploit the wider design space exposed by considering low-precision weights, activations, and accumulators, we target model deployment on FPGA accelerators with custom spatial streaming dataflow rather than general-purposes platforms like CPUs or GPUs.
The flexibility of FPGAs makes them ideal devices for low-precision inference engines as they allow for bit-level control over every part of a network; the precision of weights, activations, and accumulators can be individually tuned to custom data types for each layer without being restricted to power-of-2 bit widths like a CPU or a GPU would be.

The contributions of our work are summarized as follows:
\begin{itemize}[noitemsep,nolistsep]
\item We show that reducing the bit width of the accumulator can reduce the resource utilization of custom low-precision QNN inference accelerators.
\item We derive comprehensive bounds on accumulator bit widths with finer granularity than existing literature.
\item We introduce a novel quantization-aware training (QAT) algorithm that constrains learned parameters to avoid numerical overflow when reducing the precision of accumulators during inference.
\item We show that our algorithm not only constrains weights to fit into an accumulator of user-defined bit width, but also significantly increases the sparsity and compressibility of the resulting weights.
\item We integrate our algorithm into the Brevitas quantization library~\cite{brevitas} and the FINN compiler~\cite{xilinx2023finn} to demonstrate an end-to-end flow for training and deploying QNNs using low-precision accumulators when targeting custom streaming architectures on AMD-Xilinx FPGAs.
\end{itemize}
To the best of our knowledge, we are the first to explore the use of low-precision accumulators to improve the design efficiency of programmable QNN inference accelerators.
However, our results have implications outside of the accelerators generated by FINN.
Constraining the accumulator bit width to a user-defined upper bound has been shown to increase throughput and bandwidth efficiency on general-purpose processors~\cite{de2020quantization,ni2021wrapnet,xie2021overflow} and reduce the compute overhead of homomorphic encryption arithmetic~\cite{lou2019she}.
Furthermore, our experiments show that our algorithm can offer a better trade-off between resource utilization and model accuracy than existing approaches, confirming the benefit of including the accumulator bit width in the overall hardware-software (HW) co-design space.

\section{Related Work}
\label{sec:related_work}

\begin{figure}[t!]
\centering
\includegraphics[width=0.8\linewidth]{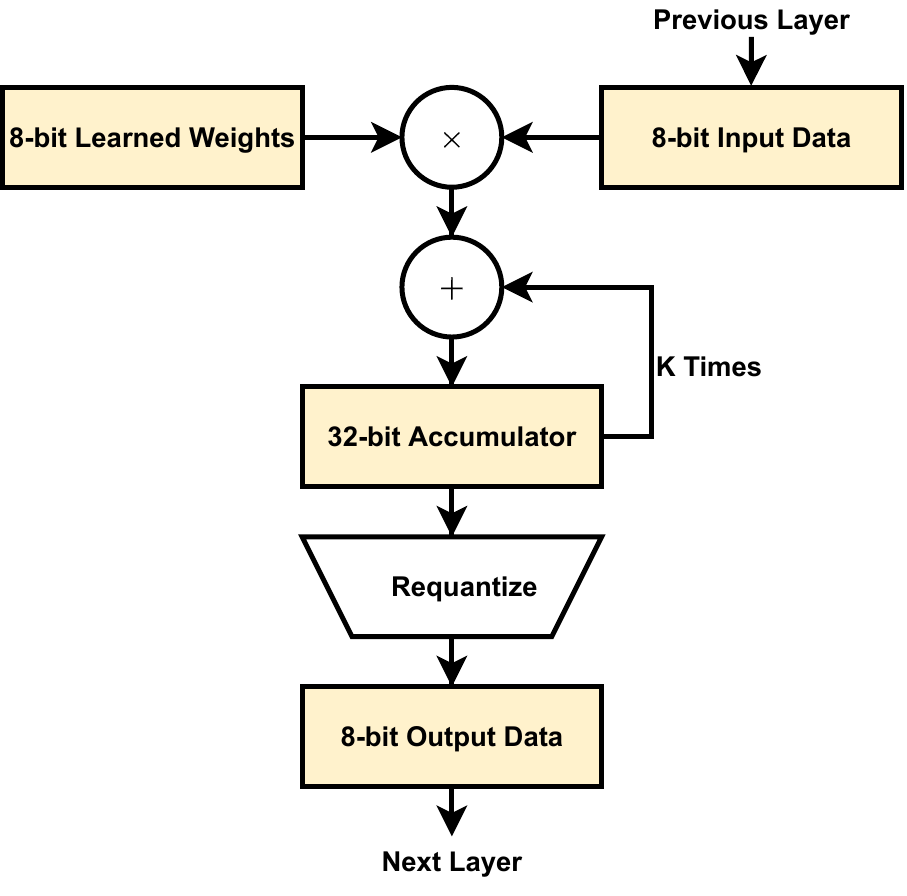}
\caption{A simplified illustration of fixed-point arithmetic in neural network inference. Quantized weights are frozen during inference. Input/output data is dynamic and thus, scaled then clipped as the hidden representations (\textit{i.e.}, activations) are passed through the network. The accumulator needs to be big enough to fit the dot product of the learned weights with input data vectors, which are assumed to both be $K$-dimensional.}
\label{fig:accumulator_flow_diagram}
\end{figure}

As activations propagate through the layers of a QNN, the intermediate partial sums resulting from convolutions and matrix multiplications are typically accumulated in a high-precision register before being requantized and passed to the next layer, which we depict in Fig.~\ref{fig:accumulator_flow_diagram}.
While many researchers and practitioners have taken to leveraging reduced precision representations for weights and activations~\cite{jacob2018quantization, gholami2021survey, nagel2021white, zhang2022learning}, few works have focused attention on reduced precision accumulators~\cite{sakr2019accumulation, de2020quantization, xie2021overflow, ni2021wrapnet}.

\begin{figure*}[t!]
\subfloat[FINN Framework Overview]{\includegraphics[width=0.6\linewidth]{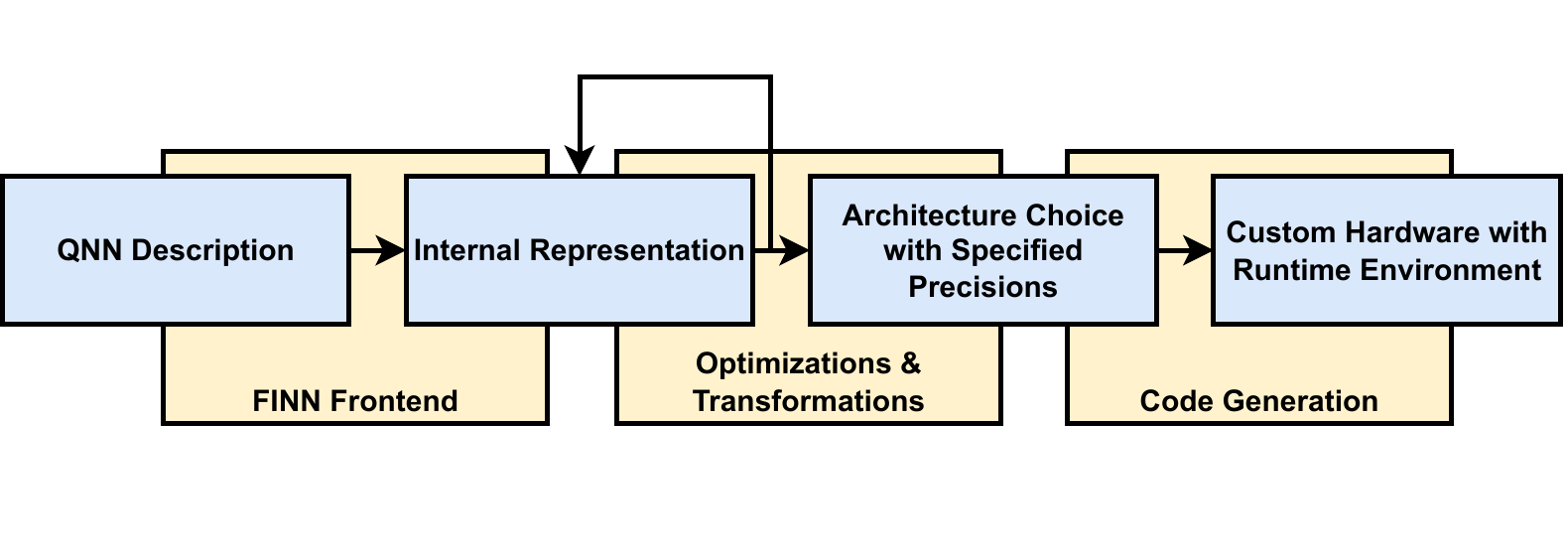}}
\subfloat[MVAU]{\includegraphics[width=0.4\linewidth]{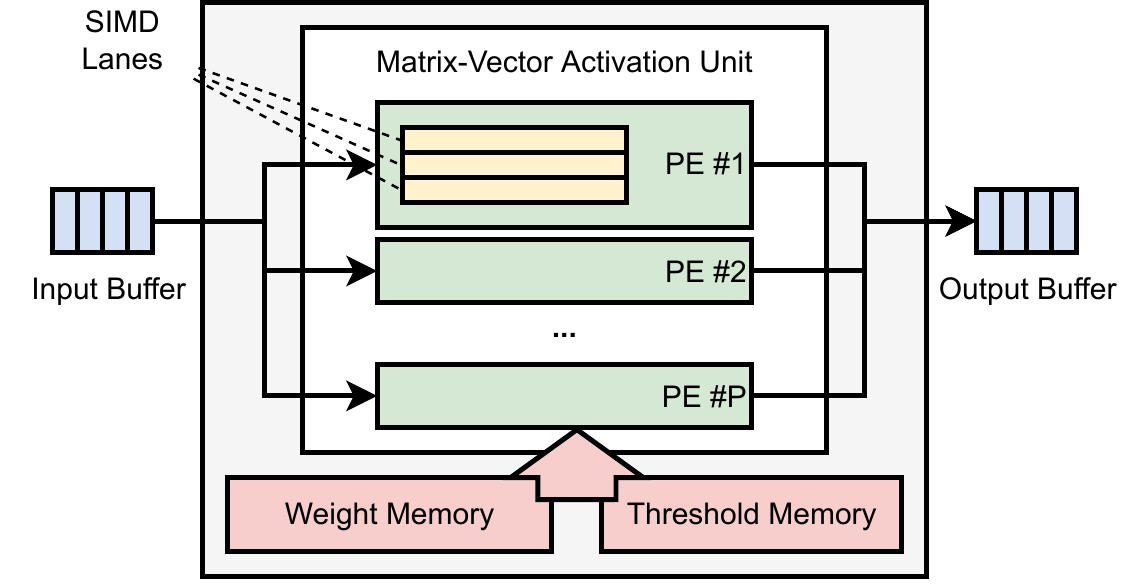}}
\caption{We adapt images from~\cite{finn, blott2018finn} to provide: (a) an overview of the FINN framework; and (b) an abstraction of the matrix-vector-activation unit (MVAU), which is one of the primary building blocks used by the FINN compiler to generate custom streaming architectures.}
\label{fig:finn}
\end{figure*}

One approach to training QNNs to use low-precision accumulators is to mitigate the impact of overflow on model accuracy.
\citealt{xie2021overflow} sought to reduce the risk of overflow using an adaptive scaling factor tuned during training; however, their approach relies on distributional assumptions that cannot guarantee overflow avoidance during inference.
Alternatively, \citealt{ni2021wrapnet} proposed training QNNs to be robust to overflow using a cyclic activation function based on expensive modulo arithmetic.
They also use a regularization penalty to control for the amount of overflows.
In both approaches, overflow is accounted for at the outer-most level, which fails to consider possible overflow when accumulating intermediate partial sums. 
Moreover, modeling overflow at the inner-most accumulation level during QAT is not easily supported by off-the-shelf deep learning frameworks as it is not directly compatible with fake-quantization over pre-existing floating-point backends.
As such, the current practice is to either use high-precision registers or simply saturate values as they are accumulated; however, such clipping can still: (1) introduce errors that cascade when propagated through a QNN; and (2) require saturation logic, which can break associativity and add to latency and area requirements~\cite{xilinx2023saturation}.
Thus, in our work, we train QNNs to completely avoid overflow rather than simply reducing its impact on model accuracy.
 
Most similar to our work is that of~\citealt{de2020quantization}, which proposed an iterative layer-wise optimization strategy to select mixed-precision bit widths to avoid overflow using computationally expensive heuristics that assume signed bit widths for all input data types.
Our proposed method constrains weights to avoid numerical overflow through the construction of our weight normalization-based quantization formulation, which accounts for both signed and unsigned input data types while adding negligible training overhead.

Tangential to our work,~\citealt{wang2018training} and~\citealt{sakr2019accumulation} study the impact of reduced precision floating-point accumulators for the purpose of accelerating training.
Such methods do not directly translate to fixed-point arithmetic, which is the focus of this work.

\section{Background}
\label{sec:background}

Our work explores the use of weight normalization as a means of constraining weights during QAT for the purpose of avoiding overflow when using low-precision accumulators.
Here, we provide background related to this objective.

\subsection{Quantization-Aware Training (QAT)}
\label{sec:prelim_quantization}

The standard operators used to emulate quantization during training rely on uniform affine mappings from a high-precision real number to a low-precision quantized number, allowing for the core computations to use integer-only arithmetic~\cite{jacob2018quantization}.
The quantizer (Eq.~\ref{eq:quantizer}) and dequantizer (Eq.~\ref{eq:dequantizer}) are parameterized by zero-point $z$ and scaling factor $s$.
Here, $z$ is an integer value that maps to the real zero such that the real zero is exactly represented in the quantized domain, and $s$ is a strictly positive real scalar that corresponds to the resolution of the quantization function.
Scaled values are rounded to the nearest integers using half-way rounding, denoted by $\lfloor \cdot \rceil$, and elements that exceed the largest supported values in the quantized domain are clipped: $\textrm{clip}(x; n,p) = \min( \max(x ;n) ;p)$, where $n$ and $p$ are dependent on the data type of $x$.
For signed integers of bit width $b$, we assume $n=-2^{b-1}$ and $p=2^{b-1} - 1$ and assume $n=0$ and $p=2^b - 1$ when unsigned.
\begin{align}
\textrm{quantize}(x;s,z) & := \textrm{clip}(\left\lfloor \frac{x}{s} \right\rceil + z; n, p) \label{eq:quantizer} \\
\textrm{dequantize}(x;s,z) & := s \cdot (x - z) \label{eq:dequantizer}
\end{align}
It has become increasingly common to use unique scaling factors for each of the output channels of the learned weights to adjust for varied dynamic ranges~\cite{nagel2019data}.
However, extending this strategy to the activations incurs additional overhead as it requires either storing partial sums or introducing additional control logic.
As such, it is standard practice to use per-tensor scaling factors for activations and per-channel scaling factors on only the weights.
It is also common to constrain the weight quantization scheme such that $z=0$, which is referred to as symmetric quantization.
Eliminating these zero points reduces the computational overhead of cross-terms when executing inference using integer-only arithmetic~\cite{jain2020trained}.
During training, the straight-through estimator (STE)~\cite{bengio2013estimating} is used to allow local gradients to permeate the rounding function such that $\nabla_x \lfloor x \rceil = 1$ everywhere, where $\nabla_x$ denotes the local gradient with respect to $x$.

\vspace{-0.1cm}
\subsection{Weight Normalization}
\label{sec:prelim_weight_norm}

Weight normalization reparameterizes each weight vector $\bm{w}$ in terms of a parameter vector $\bm{v}$ and a scalar parameter $g$ as given in Eq.~\ref{eq:standard_weight_norm}, where $\norm{\vec{v}}_2$ is the Euclidean norm of the $K$-dimensional vector $\bm{v}$~\cite{salimans2016weight}.
This simple reparameterization fixes the Euclidean norm of weight vector $\bm{w}$ such that $\norm{\bm{w}}_2 = g$, which enables the magnitude and direction to be independently learned.
\begin{equation}
	\bm{w} = g \cdot \frac{\vec{v}}{\twoNorm{v}}
	\label{eq:standard_weight_norm}
\end{equation}
Tangential to our work, prior research has sought to leverage weight normalization as a means of regularizing long-tail weight distributions during QAT~\cite{cai2019weight}.
They replace the standard $\ell_2$-norm with an $\ell_\infty$-norm and derive a projection operator to map real values into the quantized domain.
In our work, we replace the $\ell_2$-norm with an $\ell_1$-norm to use the weight normalization parameterization as a means of constraining learned weights during training to use a pre-defined accumulator bit width during inference.

\begin{figure*}[t!]
\centering
\includegraphics[width=\linewidth]{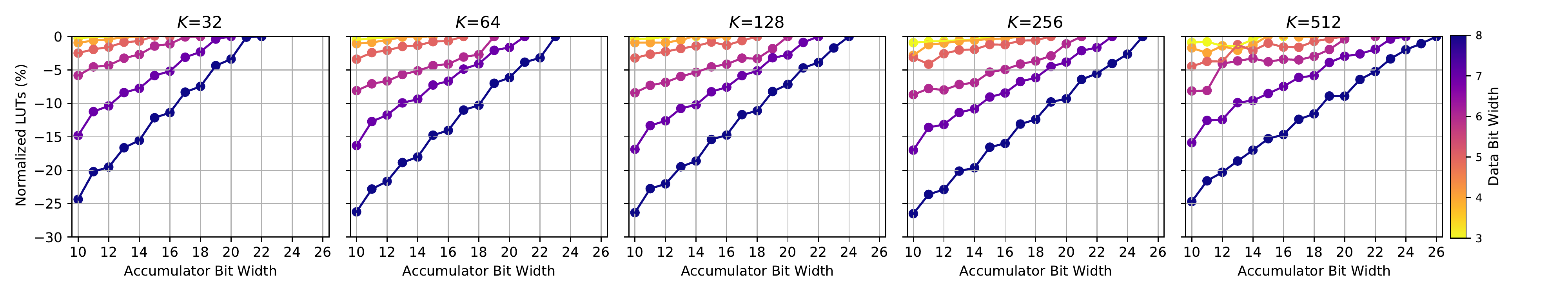}
\caption{Reducing the size of the accumulator in turn reduces LUT utilization as we vary the size of the dot product ($K$) and the input and weights bit widths, $N$ and $M$ respectively.
For simplicity, we use the same bit width for the weights and activations such that $N = M$ for all data points, and jointly refer to them as \quotes{data bit width.}
For a given dot product size and data bit width, we normalize the LUT utilization to the largest lower bound on the accumulator bit width as determined by the data types of the inputs and weights.}
\label{fig:hls_synth_unroll_1}
\end{figure*}

\vspace{-0.1cm}
\section{Motivation}
\label{sec:motivation}

To motivate our research objective, we evaluate the impact of accumulator bit width on the resource utilization of custom FPGA accelerators with spatial dataflow  architectures.
To do so, we adopt FINN~\cite{finn,blott2018finn}, an open-source framework designed to generate specialized streaming architectures for QNN inference acceleration on AMD-Xilinx FPGAs.

\subsection{Generating Streaming Architectures with FINN}
\label{sec:finn}

The FINN framework, depicted in Fig.~\ref{fig:finn}a, generates specialized QNN accelerators for AMD-Xilinx FPGAs using spatial streaming dataflow architectures that are individually customized for the network topology and the data types used.
At the core of FINN is its compiler, which empowers flexible hardware-software (HW-SW) co-design by allowing a user to have per-layer control over the generated accelerator.
Weight and activation precisions can be individually specified for each layer in a QNN,
and each layer is instantiated as its own dedicated compute unit (CU) that can be independently optimized with fine-grained parallelism.

As an example of how a layer is instantiated as its own CU, we provide a simplified abstraction of the matrix-vector-activation unit (MVAU) in Fig.~\ref{fig:finn}b.
The MVAU is one of the primary building blocks used by the FINN compiler for linear and convolutional layers~\cite{blott2018finn}.
Each CU consists of processing elements (PEs), which parallelize work along the data-independent output dimension, and single-instruction multiple-data (SIMD) lanes, which parallelize work along the data-dependent input dimension.
Execution over SIMDs and PEs within a layer is concurrent (\textit{i.e.}, spatial parallelism), while execution over layers within a network is pipelined (\textit{i.e.}, temporal parallelism).
All quantized monotonic activation functions in the network are implemented as threshold comparisons that map high-precision accumulated results from the preceding layer into low-precision output values.
During compilation, batch normalization, biases and even scaling factors are absorbed into this threshold logic via mathematical manipulation~\cite{blott2018finn}.
The input and output data for the generated accelerators are streamed into and out of the chip using AXI-Stream protocols while on-chip data streams are used to interconnect these CUs to propagate intermediate activations through the layers of the network.
During inference, all network parameters are stored on-chip to avoid external memory bottlenecks.
For more information on the FINN framework, we refer the interested reader to~\cite{finn,blott2018finn, xilinx2023finn}.

\subsection{Accumulator Impact on Resource Utilization}
\label{sec:impact}

FINN typically relies on look-up tables (LUTs) to perform MACs at low precision; in such scenarios, LUTs are often the resource bottleneck for the low-precision streaming accelerators it generates.
Furthermore, because activation functions are implemented as threshold comparisons, their resource utilization exponentially grows with the precision of the accumulator and output activations~\cite{blott2018finn}.
Thus, reducing the size of the accumulator has a direct influence on both the compute and memory requirements.

To evaluate the impact of accumulator bit width on LUT utilization, we consider a fully connected QNN with one hidden layer that is parameterized by a matrix of signed integers.
The QNN takes as input a $K$-dimensional vector of unsigned integers and gives as output a 10-dimensional vector of signed integers.
We use the FINN compiler to generate a streaming architecture with a single MVAU targeting an AMD-Xilinx PYNQ-Z2 board with a frequency of 100 MHz. We report the resource utilization of the resulting RTL post-synthesis.
To simplify our analysis, we assume that LUTs are the only type of resources available and configure the FINN compiler to target LUTs for both compute and memory so that we can evaluate the impact of accumulator bit width on resource utilization using just one resource.

As further discussed in Section~\ref{sec:bounds}, the minimum accumulator bit width that can be used to avoid overflow is a function of the size of the dot product ($K$) as well as the bit widths of the input and weight vectors $\vec{x}$ and $\vec{w}$, respectively denoted as $N$ and $M$.
In Fig.~\ref{fig:hls_synth_unroll_1}, we visualize how further reducing the accumulator bit width in turn decreases resource utilization as we vary $K$, $N$, and $M$.
For a given dot product size and data bit width, we normalize the LUT utilization to the largest lower bound on the accumulator bit width as determined by the data types of the inputs and weights.
To control for the resource utilization of dataflow logic, we use a single PE without applying optimizations such as loop unrolling, which increase the amount of SIMD lanes.

We observe that the impact of accumulator bit width on resource utilization grows exponentially with the precision of the data (\textit{i.e.}, $M$ and $N$).
As we reduce the size of the accumulator, we observe up to a 25\% reduction in the LUT utilization of a layer when $N = M = 8$, but only up to a 1\% reduction in LUT utilization when $N = M = 3$.
This is expected as compute and memory requirements exponentially grow with precision and thus have larger proportional savings opportunities.
We also observe that $K$ has a dampening effect on the impact of accumulator bit width reductions that is also proportional to the precision of the data.
When $N = M = 8$ and $K=32$, we observe on average a 2.1\% LUT reduction for every bit that we reduce the accumulator, but only a 1.5\% LUT reduction when $K=512$.
Conversely, we observe on average a 0.2\% LUT reduction for every bit that we reduce the accumulator when $N = M = 3$ regardless of $K$.
We hypothesize that this dampening effect is in part due to the increased storage costs of larger weight matrices because, unlike threshold storage, the memory requirements of weights are not directly impacted by the accumulator bit width.
Therefore, the relative savings from accumulator bit width reductions are diluted by the constant memory requirements of the weights.
We explore this further in Section~\ref{sec:experiments}, where we break down the resource utilization of FPGA accelerators generated for our benchmark models.

\section{Accumulator Bit Width Bounds}
\label{sec:bounds}

Figure~\ref{fig:accumulator_flow_diagram} illustrates a simplified abstraction of accumulation in QNN inference.
As activations are propagated through the layers, the intermediate partial sums resulting from operations such as convolutions or matrix multiplications are accumulated into a register before being requantized and passed to the next layer.
To avoid numerical overflow, the register storing these accumulated values needs to be wide enough to not only store the result of the dot product, but also all intermediate partial sums.

Consider the dot product of input data $\vec{x}$ and learned weights $\vec{w}$, which are each $K$-dimensional vectors of integers.
Let $y$ be the scalar result of their dot product given by Eq.~\ref{eq:dot_product}, where $x_i$ and $w_i$ denote element $i$ of vectors $\bm{x}$ and $\bm{w}$, respectively.
Since the representation range of $y$ is bounded by that of $\vec{x}$ and $\vec{w}$, we use their ranges to derive lower bounds on the bit width $P$ of the accumulation register, or accumulator.
\begin{equation}
y = \textstyle \sum_{i=1}^K x_i w_i
\label{eq:dot_product}
\end{equation}
It is common for input data to be represented with unsigned integers either when following activation functions with non-negative dynamic ranges (\textit{e.g.}, rectified linear units, or ReLUs), or when an appropriate zero point is adopted (\textit{i.e.}, asymmetric quantization).
Otherwise, signed integers are used.
Since weights are most often represented with signed integers, we assume the accumulator is always signed in our work.
Therefore, given that the scalar result of the dot product between $\vec{x}$ and $\vec{w}$ is a $P$-bit integer defined by Eq.~\ref{eq:dot_product}, it follows that $\textstyle \sum_{i=1}^K x_i w_i$ is bounded such that:
\begin{equation}
-2^{P - 1} \leq \textstyle \sum_{i=1}^K x_i w_i \leq 2^{P - 1} - 1
\label{eq:basic_bound_on_dot_prod}
\end{equation}
To satisfy the right-hand side of this double inequality, it follows that $\vert \textstyle \sum_{i=1}^K x_i w_i \vert \leq 2^{P - 1} - 1$.
However, the accumulator needs to be wide enough to not only store the result of the dot product, but also all intermediate partial sums.

Since input data is not known \textit{a priori}, our bounds must consider the worst-case values for every MAC.
Thus, because the magnitude of the sum of products is upper-bounded by the sum of the product of magnitudes, it follows that if $\textstyle \sum_{i=1}^K \vert x_i \vert \vert w_i \vert \leq 2^{P - 1} - 1$, then the dot product between $\vec{x}$ and $\vec{w}$ fits into a $P$-bit accumulator without numerical overflow, as shown below.
\begin{equation}
\vert \textstyle \sum_i x_i w_i \vert \leq \textstyle \sum_i \vert x_i w_i \vert \leq \textstyle \sum_i \vert x_i \vert \vert w_i \vert \leq 2^{P - 1} - 1
\label{eq:mag_identities}
\end{equation}

\subsection{Deriving Lower Bounds Using Data Types}
\label{sec:datatype_bounds}

The worst-case values for each MAC can na\"ively be inferred from the representation range of the data types used.
When $x_i$ and $w_i$ are signed integers, their magnitudes are bounded such that $\vert x_i \vert \leq 2^{N - 1}$ and $\vert w_i \vert \leq 2^{M - 1}$, respectively.
In scenarios where $x_i$ is an unsigned integer, the magnitude of each input value is upper-bounded such that $\vert x_i \vert \leq 2^N - 1$; however, we simplify this upper bound to be $\vert x_i \vert \leq 2^N$ for convenience of notation\footnote{
Note that our simplification of the upper bound for unsigned input data does not compromise overflow avoidance.}.
Combining the signed and unsigned upper bounds, it follows that $\vert x_i \vert \leq 2^{N - \mathbbm{1}_\text{signed}(\vec{x})}$, where $\mathbbm{1}_\text{signed}(\bm{x})$ is an indicator function that returns 1 if and only if $\bm{x}$ is a vector of signed integers.

Building from Eq.~\ref{eq:mag_identities}, it follows that the sum of the product of the magnitudes is bounded such that:
\begin{equation}
\textstyle  \sum_{i=1}^K \vert x_i \vert \vert w_i \vert \leq K \cdot 2^{N + M - 1 - \mathbbm{1}_\text{signed}(\bm{x})} \leq 2^{P - 1} - 1
\label{eq:unfinished_datatype_bound}
\end{equation}
Taking the log of both sides of  Eq.~\ref{eq:unfinished_datatype_bound}, we can derive a lower bound on the accumulator bit width $P$:
\begin{equation}
\log_2 \left(2^{\log_2(K) + N + M - 1 - \mathbbm{1}_\text{signed}(\bm{x})} + 1 \right) + 1 \leq P
\end{equation}
This simplifies to the following lower bound on $P$:
\begin{align}
P & \geq \alpha + \phi(\alpha) + 1 \label{eq:datatype_lower_bound}\\
\alpha & = \log_2(K) + N + M - 1 - \mathbbm{1}_\text{signed}(\bm{x}) \label{eq:datatype_alpha}\\
\phi(\alpha) & = \log_2(1 + 2^{-\alpha}) \label{eq:phi_alpha}
\end{align}
In Fig.~\ref{fig:acc_bit_width_bounds}, we visualize this bound assuming that $\vec{x}$ is a vector of unsigned integers such that $\mathbbm{1}_\text{signed}(\bm{x})=0$.
There, we show how the lower bound on the accumulator bit width increases as we vary both the size of the dot product ($K$) and the bit width of the weights and activations.

\subsection{Deriving Lower Bounds Using Learned Weights}
\label{sec:weight_bounds}

Since learned weights are frozen during inference time, we can use knowledge of their magnitudes to derive a tighter lower bound on the accumulator bit width.

\begin{figure}[t!]
\centering
\includegraphics[width=\linewidth]{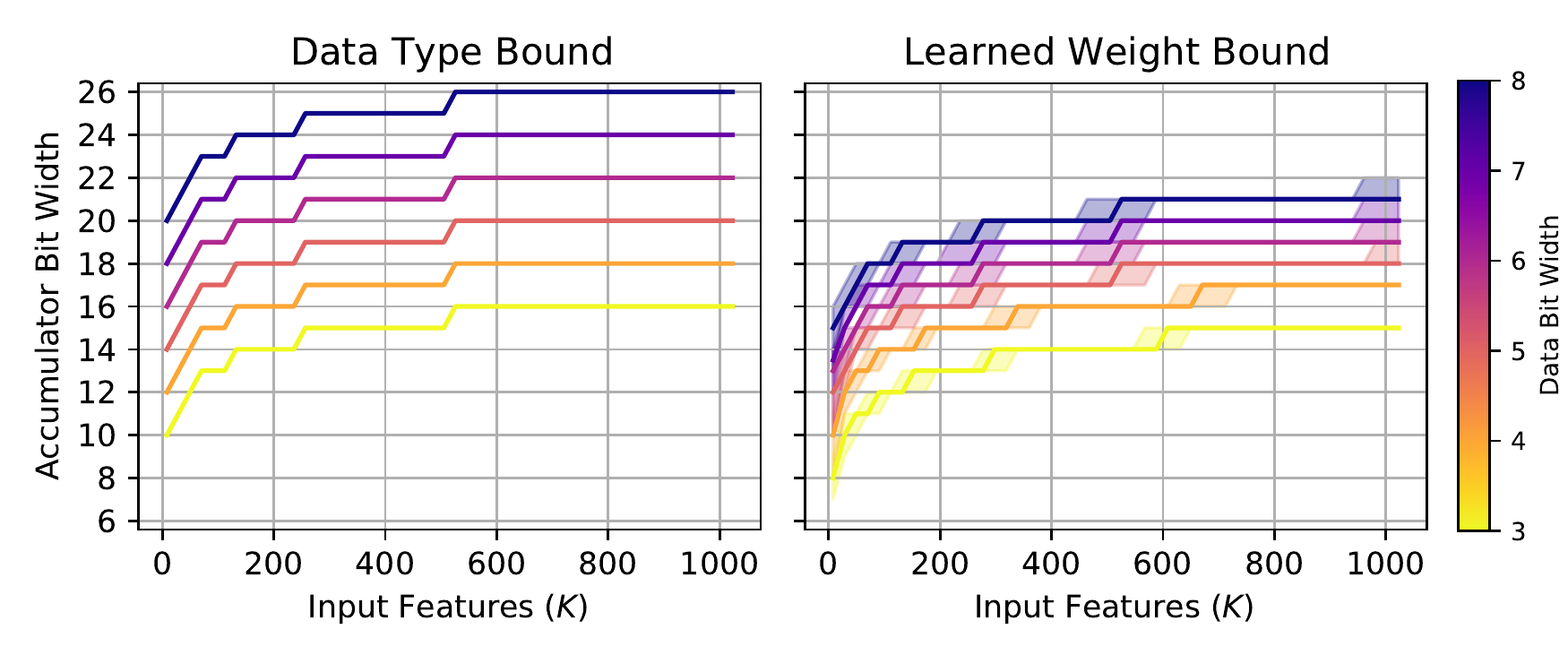}
\caption{We visualize the differences between our accumulator bit width bounds as we vary the size of the dot product ($K$) as well as the bit width of the weights ($M$) and input activations ($N$), which we jointly refer to as \quotes{data bit width} such that $M=N$.}
\label{fig:acc_bit_width_bounds}
\end{figure}

Building again from Eq.~\ref{eq:mag_identities}, the sum of the product of magnitudes is bounded by Eq.~\ref{eq:first_weight_bound_eq}, where $\Vert \bm{w} \Vert_1$ denotes the standard $\ell_1$-norm over vector $\bm{w}$.
\begin{equation}
\textstyle \sum_{i=1}^K \vert x_i \vert \vert w_i \vert \leq 2^{N - \mathbbm{1}_\text{signed}(\bm{x})} \cdot \Vert \bm{w} \Vert_1 \leq 2^{P - 1} - 1
\label{eq:first_weight_bound_eq}
\end{equation}
Here, we define a tighter lower bound on $P$:
\begin{align}
P & \geq \beta + \phi(\beta) + 1\\
\beta &=  \log_2(\Vert \bm{w} \Vert_1) + N - \mathbbm{1}_\text{signed}(\bm{x}) \label{eq:weight_bound}\\
\phi(\beta) & = \log_2(1 + 2^{-\beta})
\end{align}
In Fig.~\ref{fig:acc_bit_width_bounds}, we visualize this bound again assuming that $\vec{x}$ is a vector of unsigned integers.
Because Eq.~\ref{eq:weight_bound} is dependent on the values of the learned weights, we randomly sample each $K$-dimensional vector from a discrete Gaussian distribution and show the median accumulator bit width along with the minimum and maximum observed over 300 random samples.
We show that using learned weights (right) provides a tighter lower bound on the bit width of the accumulator than using data types (left) as we vary both the size of the dot product ($K$) and the bit width of the weights and input activations.

\section{Training QNNs to Avoid Overflow}
\label{sec:method}

To train QNNs to use low-precision accumulators without overflow, we use weight normalization as a means of constraining learned weights $\vec{w}$ to satisfy the bound derived in Section~\ref{sec:weight_bounds}.
Building from Eq.~\ref{eq:first_weight_bound_eq}, we transform our lower bound on accumulator bit width $P$ to be the upper bound on the $\ell_1$-norm of $\vec{w}$ given by Eq.~\ref{eq:weight_bound_reparameterized}.
Note that because each output neuron requires its own accumulator, this upper bound needs to be enforced channelwise.
\begin{equation}
\Vert \bm{w} \Vert_1 \leq \left( 2^{P - 1} - 1 \right) \cdot 2^{\mathbbm{1}_\text{signed}(\bm{x}) - N}
\label{eq:weight_bound_reparameterized}
\end{equation}

\subsection{Constructing Our Quantization Operator}

To enforce this constraint during QAT, we reparameterize our quantizer such that each weight vector $\vec{w}$ is represented in terms of parameter vectors $\vec{g}$ and $\vec{v}$.
Similar to the standard weight normalization formulation discussed in Section~\ref{sec:prelim_weight_norm}, this reparameterization decouples the norm from the weight vector; however, unlike the standard formulation, the norm is learned for each output channel.
For a given layer with $C$ output channels, we replace the per-tensor $\ell_2$-norm of the standard formulation (Eq.~\ref{eq:standard_weight_norm}) with a per-channel $\ell_1$-norm.
This reparameterization, given by Eq.~\ref{eq:our_weight_normalization}, allows for the $\ell_1$-norm of weight vector $\vec{w}$ to be independently learned per-channel such that $g_i = \Vert \bm{w}_i \Vert_1$ for all $i \in \{1, \cdots, C \}$, where $\vec{w}_i$ denotes the weights of channel $i$ and $g_i$ denotes element $i$ in parameter vector $\vec{g}$.
\begin{equation}
\vec{w}_i = g_i \cdot \frac{\vec{v_i}}{\oneNorm{v_i}}	\quad \forall ~i \in \{1, \cdots, C \}
\label{eq:our_weight_normalization}
\end{equation}
Similar to the standard quantization operator, our weight normalization-based quantization relies on a uniform affine mapping from the high-precision real domain to the low-precision quantized domain using learned per-channel scaling factors $\vec{s}=\{s_i\}^C_{i=1}$.
Thus, by constraining $g_i$ to satisfy Eq.~\ref{eq:constrain_g}, 
we can learn quantized weights that satisfy our accumulator bit width bound and avoid overflow.
\begin{equation}
g_i \leq s_i \cdot \left( 2^{P - 1} - 1 \right) \cdot 2^{\mathbbm{1}_\text{signed}(\bm{x}) - N}
\label{eq:constrain_g}
\end{equation}
Below, we articulate our weight normalization-based quantization operator.
For clarity and convenience of notation, we consider a layer with one output channel (\textit{i.e.}, $C=1$) such that parameter vectors $\vec{g}=\{g_i\}^C_{i=1}$ and $\vec{s}=\{s_i\}^C_{i=1}$ can be represented as scalars $g$ and $s$, respectively.
\begin{equation}
\textrm{quantize}(\bm{w};s,z) := \textrm{clip}(\left\lfloor \frac{g}{s} \frac{\vec{v}}{\oneNorm{v}} \right\rfloor + z; n, p) \label{eq:quantizer_ours}
\end{equation}
During training, our weight quantization operator applies fours elementwise operations the following in order: scale, round, clip, then dequantize.
As with the standard operator, we eliminate the zero points in our mapping such that $z=0$.
We use an exponential parameterization of both the scaling factor $s = 2^{d}$ and the norm parameter $g = 2^t$, where $d$ and $t$ are both log-scale parameters to be learned through stochastic gradient descent.
This is similar to the work of~\cite{jain2020trained} with the caveat that we remove integer power-of-2 constraints, which provide no added benefit to the streaming architectures generated by FINN as floating-point scaling factors can be absorbed into the threshold logic via mathematical manipulation~\cite{blott2018finn}.
The scaled tensors are then rounded to zero, which we denote by $\lfloor \cdot \rfloor$.
This prevents any upward rounding that may cause the norm to increase past our constraint.
Note that this is another difference from the conventional quantization operators, which use half-way rounding.
Finally, once scaled and rounded, the elements in the tensor are then clipped and dequantized using Eq.~\ref{eq:dequantizer}.
Our resulting quantization operator used during training is given by Eq.~\ref{eq:qat_weight_norm}, where $n$ and $p$ depend on the representation range of weight bit width $M$.

\begin{align}
q(\bm{w} ; s) & := \text{clip}\left( \left\lfloor \frac{g}{s} \frac{\vec{v}}{\oneNorm{v}} \right\rfloor ; n, p \right) \cdot s \label{eq:qat_weight_norm} \\
\text{ where } & s = 2^{d} \\
\text{ and } & g = 2^{\min(T, t)} \label{eq:min_g_t_T} \\
\text{ and } & T = \mathbbm{1}_\text{signed}(\bm{x}) + \log_2(2^{P - 1} - 1) + d - N \label{eq:T}
\end{align}

When quantizing our activations, we use the standard quantization operators discussed in Section~\ref{sec:prelim_quantization}.
All activations that follow non-negative functions (\textit{i.e.}, ReLU) are represented using unsigned integers, otherwise they are signed.

To update our learnable parameters during training, we use the straight-through estimator (STE)~\cite{bengio2013estimating} to allow local gradients to permeate our rounding function such that $\nabla_x \left\lfloor x \right\rfloor = 1$ everywhere, as is common practice.

\begin{figure*}[t!]
\centering
\includegraphics[width=\linewidth]{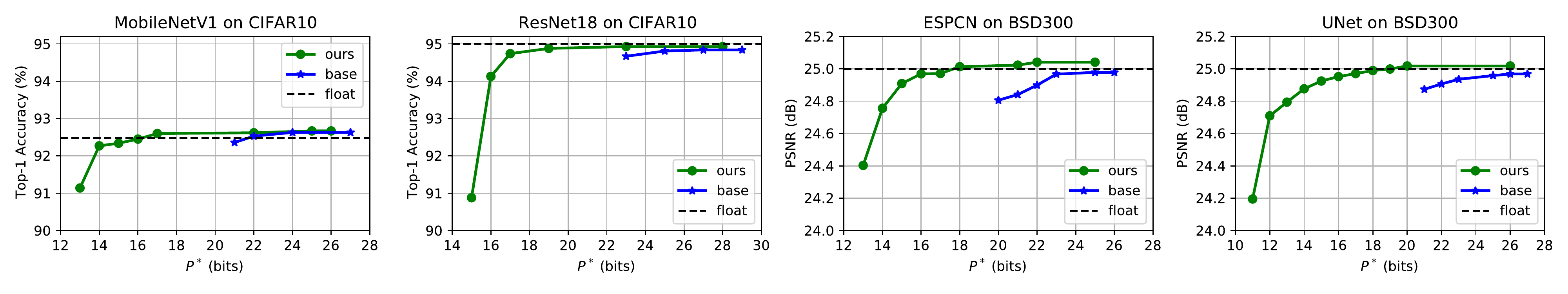}
\caption{Using image classification and single-image super resolution benchmarks, we show that we are able to maintain model performance with respect to the floating-point baseline even with significant reductions to the accumulator bit width.
We visualize this trade-off using pareto frontiers estimated using a grid search over various weight, activation, and accumulator bit widths.
Here, we use $P^*$ to denote the largest accumulator bit width allowed across all layers in the network.
We compare our algorithm (\textcolor{ForestGreen}{\textbf{green dots}}) against the standard quantization baseline algorithm (\textcolor{blue}{\textbf{blue stars}}) and repeat each experiment 3 times, totaling over 500 runs per model to form each set of pareto frontiers.
We observe that our algorithm dominates the baseline in all benchmarks, showing that we can reduce the accumulator bit width without sacrificing significant model performance even with respect to a floating-point baseline.
}
\label{fig:pareto_fronts_adt}
\end{figure*}

\subsection{Regularization with Lagrangian Penalties}

To avoid $t$ getting stuck when $t > T$ in Eq.~\ref{eq:min_g_t_T}, we introduce the Lagrangian penalty $\mathcal{L}_\text{penalty}$ given by Eq.~\ref{eq:penalty}.
For a neural network with $L$ layers, each with $C_l$ output channels, $t_{i,l}$ denotes the log-scale parameter of the norm for channel $i$ in layer $l$ and $T_{i,l}$ denotes its upper bound as given by Eq.~\ref{eq:T}.
Note that, even when $\mathcal{L}_\text{penalty} > 0$, we still satisfy our accumulator constraints by clipping the norm in Eq.~\ref{eq:min_g_t_T}.
However, our Lagrangian penalty encourages norm $g_i$ and scale $s_i$ of each channel $i$ in each layer to jointly satisfy the bound without clipping, which allows their log-scale parameters to be updated with respect to only the task error.

\begin{equation}
\mathcal{L}_\text{penalty} = \textstyle \textstyle \sum_{l=1}^L \sum_{i=1}^{C_l} \left(t_{i,l} - T_{i,l} \right)_+
\label{eq:penalty}
\end{equation}

It is important to note that our formulation is task agnostic.
Assuming base task loss $\mathcal{L}_\text{task}$, our total loss $\mathcal{L}_\text{total}$ is given by Eq.~\ref{eq:lagrangians}, where $\lambda$ is a Lagrange multiplier.
In our experiments, we fix $\lambda$ to be a constant scalar, although adaptive approaches such as~\cite{roy2021direct} could be explored.
\begin{align}
\mathcal{L}_\text{total} & = \mathcal{L}_\text{task} + \lambda \mathcal{L}_\text{penalty}  \label{eq:lagrangians}
\end{align}

\section{Experiments}
\label{sec:experiments}

We evaluate our algorithm using image classification and single-image super resolution benchmarks.
Because our algorithm is the first of its kind, we compare our approach to the standard QAT formulation discussed in Section~\ref{sec:prelim_quantization}.
We implement all algorithms in PyTorch using Brevitas v0.7.2~\cite{brevitas}, where single-GPU quantization-aware training times range from 5 minutes (\textit{e.g.}, ESPCN) to 2 hours (\textit{e.g.}, MobileNetV1) on an AMD MI100 accelerator.
To generate custom FPGA architectures for the resulting QNNs, we work from FINN v0.8.1~\cite{xilinx2023finn} and add extensions to support our work.

\subsection{Image Classification Benchmarks}
\label{sec:cifar10}

We train MobileNetV1~\cite{howard2017mobilenets} and ResNet18~\cite{he2016deep} to classify images using the CIFAR10 dataset~\cite{krizhevsky2009learning}.
We closely follow the network architectures originally proposed by the respective authors, but introduce minor variations that yield more amenable intermediate representations given the image size as we discuss below.
As is common practice, we fix the input and output layers to 8-bit weights and activations for all configurations, and initialize all models from floating-point counterparts pre-trained to convergence on CIFAR10.
We evaluate all models by the observed top-1 test accuracy.

For MobileNetV1, we use a stride of 2 for both the first convolution layer and the final average pooling layer.
This reduces the degree of downscaling to be more amenable to training over smaller images.
All other layer configurations remain the same as proposed in~\cite{howard2017mobilenets}.
We use the stochastic gradient descent (SGD) optimizer to fine-tune all models for 100 epochs in batches of 64 images using a weight decay of 1e-5.
We use an initial learning rate of 1e-3 that is reduced by a factor of 0.9 every epoch.

For ResNet18, we alter the first convolution layer to use a stride and padding of 1 with a kernel size of 3. Similar to MobileNetV1, we remove the preceding max pool layer to reduce the amount of downscaling throughout the network.
We also use a convolution shortcut~\cite{he2016identity} rather than the standard identity as it empirically proved to yield superior results in our experiments.
All other layer configurations remain the same as proposed in~\cite{he2016deep}.
We use the SGD optimizer to fine-tune all models for 100 epochs in batches of 256 using a weight decay of 1e-5.
We use an initial learning rate of 1e-3 that is reduced by a factor of 0.1 every 30 epochs.

\subsection{Single-Image Super Resolution Benchmarks}
\label{sec:bsd300}

We train ESPCN~\cite{shi2016real} and UNet~\cite{ronneberger2015u} to upscale single images by a factor of 3x using the BSD300 dataset~\cite{MartinFTM01}.
Again, we closely follow the network architectures originally proposed by the respective authors, but introduce minor variations that yield more hardware-friendly network architectures.
As is common practice, we fix the input and output layers to 8-bit weights and activations for all configurations; however, we train all super resolution models from scratch.
We empirically evaluate all models by the peak signal-to-noise ratio (PSNR) observed over the test dataset.

For ESPCN, we replace the sub-pixel convolution with a nearest neighbor resize convolution (NNRC), which has been shown to reduce checkerboard artifacts during training~\cite{odena2016deconvolution} and can be efficiently executed during inference~\cite{ colbert2021energy}.
All other layer configurations remain the same as  proposed in~\cite{shi2016real}.
We use the Adam optimizer~\cite{kingma2014adam} to fine-tune all models for 100 epochs in batches of 16 images using a weight decay of 1e-4.
We use an initial learning rate of 1e-4 that is reduced by a factor of 0.98 every epoch.

For UNet, we use only 3 encoders and decoders to create a smaller architecture than originally proposed by~\cite{ronneberger2015u}. 
We replace transposed convolutions with NNRCs, which have been shown to be functionally equivalent during inference~\cite{colbert2021energy}, but have more favorable behavior during training~\cite{odena2016deconvolution}.
We replace all concatenations with additions and reduce the input channels accordingly.
We use the Adam optimizer to fine-tune all models for 200 epochs in batches of 16 images using a weight decay of 1e-4.
We use an initial learning rate of 1e-3 that is reduced by a factor of 0.3 every 50 epochs.

\begin{figure*}[t!]
\centering
\subfloat[No Parallelism]{\includegraphics[width=\linewidth]{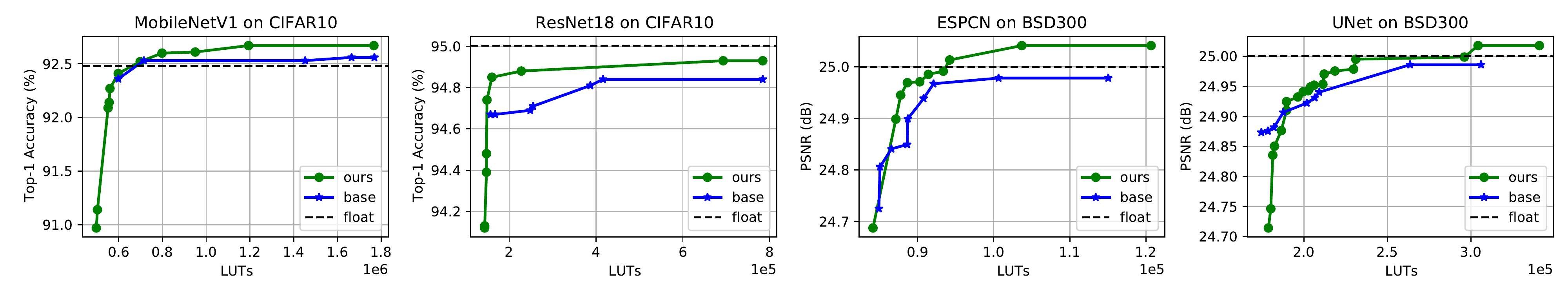}} \\
\subfloat[Max PEs]{\includegraphics[width=\linewidth]{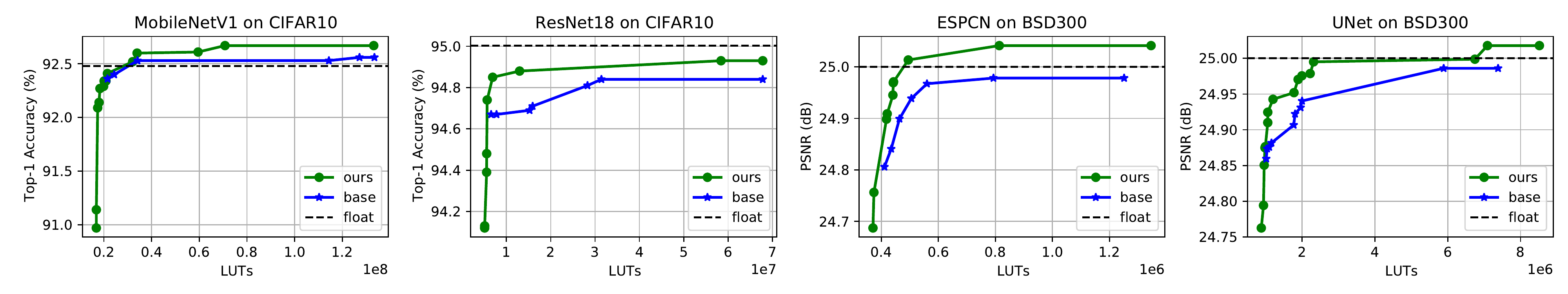}} \\
\subfloat[Max PEs and SIMDs]{\includegraphics[width=\linewidth]{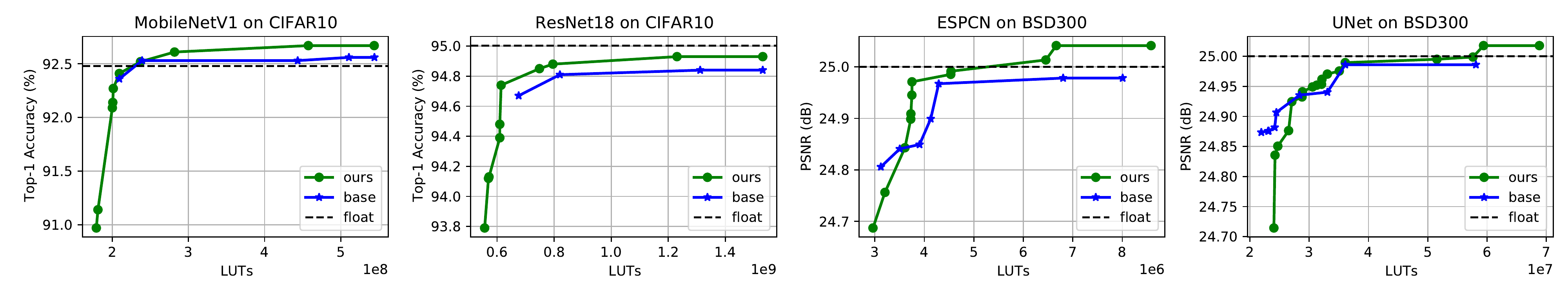}}
\caption{To evaluate the trade-off between resource utilization and accuracy, we visualize the pareto frontier observed over our image classification and single-image super resolution benchmarks.
We evaluate these pareto frontiers in the following scenarios: (a) no spatial parallelism; (b) maximizing the PEs for each layer; and (c) maximizing the PEs and SIMDs for each layer.
In each scenario, we observe that our algorithm (\textcolor{ForestGreen}{\textbf{green dots}}) provides a dominant pareto frontier across each model when compared to the standard quantization algorithm (\textcolor{blue}{\textbf{blue stars}}), showing that our algorithm can reduce LUT utilization without sacrificing significant model performance.}
\label{fig:pareto_fronts_lut}
\end{figure*}

\subsection{Experiment Setup and Research Questions}
\label{sec:exp_details}

We design our experiments around the following questions:
\vspace{-0.25cm}
\begin{itemize}[noitemsep,nolistsep]
\item How does reducing the accumulator bit width impact model performance (Section~\ref{sec:accumulator_impact})?
\item What are the trade-offs between resource utilization and model performance (Section~\ref{sec:luts_vs_perf_scatter})?
\item Where do our resource savings come from as we reduce accumulator bit width (Section~\ref{sec:resource_savings})?
\end{itemize}

Similar to the experiments described in Section~\ref{sec:motivation}, we simplify our analysis and assume that LUTs are the only type of resources available.
We configure the FINN compiler to target LUTs for both compute and memory wherever possible so that we can evaluate the impact of accumulator bit width on resource utilization using just one type of resource.

Throughout our experiments, we focus our attention on data bit widths between 5 and 8 bits for two reasons: (1) reducing precision below 5 bits often requires uniquely tailored hyperparameters, which would not make for an even comparison across bit widths; and (2) reducing the size of the accumulator has a negligible impact on LUT utilization at lower data bit widths, as shown in Section~\ref{sec:motivation}.
Still, even with a reduced set of possible data bit widths, it is computationally intractable to test every combination of weight, activation, and accumulator bit widths within the design space exposed by the QNNs that we use as benchmarks.
Thus, for weight and activation bit widths, we uniformly enforce precision for each hidden layer in the network such that $M$ and $N$ are constant scalars, aside from the first and last layers that remain at 8 bits such that $M=N=8$.
The accumulator bit width, however, is dependent on not only the weight and activation bit widths, but also the size of the dot product ($K$), as discussed in Section~\ref{sec:bounds}.
To simplify our design space, we constrain all layers in a given network to use the same accumulator bit width that we denote as $P^*$ such that the maximum accumulator bit width for any layer is $P^*$ bits.
Recall that the value for $P^*$ is used by Eq.~\ref{eq:weight_bound_reparameterized} to upper bound the $\ell_1$-norm of each weight per-channel and used by Eq.~\ref{eq:T} to enforce this constraint during training.

To collect enough data to investigate our research questions, we perform a grid search over weight and activation bit widths from 5 to 8 bits.
For each of these 16 combinations, we calculate the largest lower bound on the accumulator bit width as determined by the data type bound (Eq.~\ref{eq:datatype_lower_bound}) of the largest layer in the network.
Using this to initialize $P^*$, we evaluate up to a 10-bit reduction in the accumulator bit width to create a total of 160 configurations.
Finally, we benchmark our results against the standard quantization algorithm discussed in Section~\ref{sec:prelim_quantization}; however, because it does not expose control over accumulator bit width, this grid search is restricted to the 16 combinations of weight and activation bit widths.
We run each configuration 3 times to form a total of 528 runs per model.
In the following sections, we summarize our findings.

\begin{figure*}[t!]
\centering
\subfloat[No Paralleism]{\includegraphics[width=\linewidth]{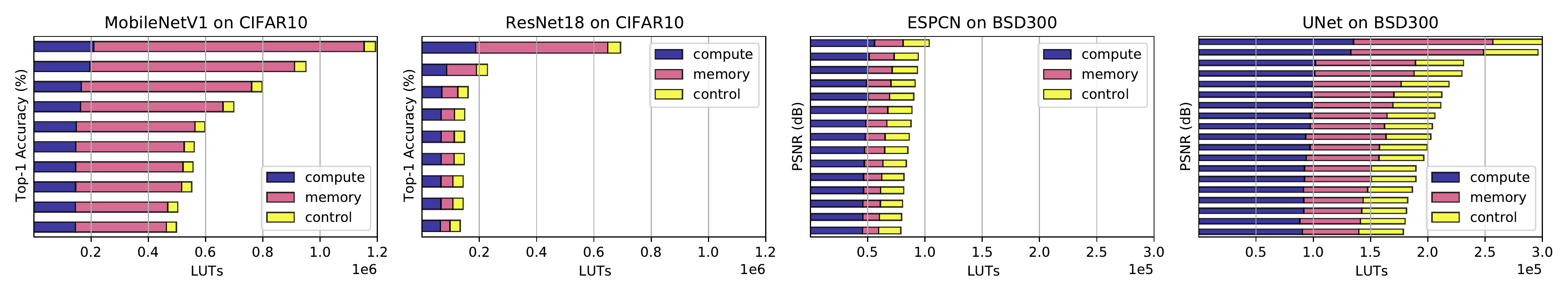}} \\
\subfloat[Max PEs]{\includegraphics[width=\linewidth]{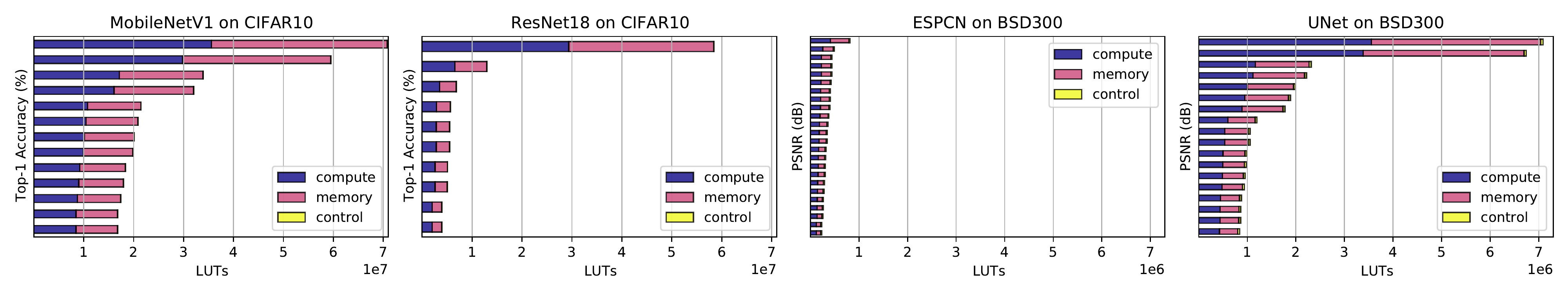}} \\
\subfloat[Max PEs and SIMDs]{\includegraphics[width=\linewidth]{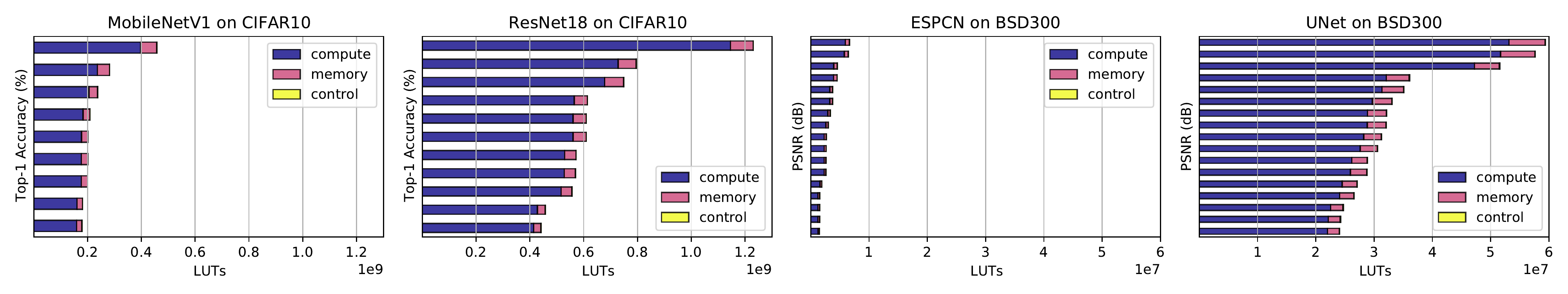}}
\caption{We break down LUT utilization into compute, memory, and control flow for each of our pareto optimal models from Fig.~\ref{fig:pareto_fronts_lut}.}
\label{fig:pareto_fronts_breakdown}
\end{figure*}

\subsection{Accumulator Impact on Model Performance}
\label{sec:accumulator_impact}

Our algorithm introduces a novel means of constraining the weights of a QNN to use a pre-defined accumulator bit width without overflow.
As an alternative to our algorithm, a designer can choose to heuristically manipulate data bit widths based on our data type bound given by Eq.~\ref{eq:datatype_lower_bound}.
Such an approach would still guarantee overflow avoidance when using a pre-defined accumulator bit width $P$, but is an indirect means of enforcing such a constraint.
Given a pre-defined accumulator bit width upper bound $P^*$, we compare the performance of models trained with our algorithm against this heuristic approach.
We visualize this comparison as a pareto frontier in Fig.~\ref{fig:pareto_fronts_adt}.
It is important to note that, while this is not a direct comparison against the algorithm proposed by~\cite{de2020quantization}, the experiment is similar in principle.
Unlike~\cite{de2020quantization}, we use the more advanced quantization techniques detailed in Section~\ref{sec:prelim_quantization}, and replace the computationally expensive loss-guided search technique with an even more expensive, but more comprehensive grid search.

The pareto frontier shows the maximum observed model performance for a given $P^*$.
We observe that our algorithm can push the accumulator bit width lower than what is attainable using current methods while also maintaining model performance.
Furthermore, most models show less than a 1\% performance drop from even the floating-point baseline with a 16-bit accumulator, which is most often the target bit width for low-precision accumulation in general-purpose processors~\cite{de2020quantization, xie2021overflow}.

\subsection{Trade-Offs Between Resources and Accuracy}
\label{sec:luts_vs_perf_scatter}

To understand the impact that accumulator bit width can have on the design space of the accelerators generated by FINN, we evaluate the trade-offs between resource utilization and model performance.
For each of the models trained in the grid search detailed in Section~\ref{sec:exp_details}, we use the FINN compiler to generate resource utilization estimates and use pareto frontiers to visualize the data.
In Fig.~\ref{fig:pareto_fronts_lut}, we provide the maximum observed model performance for the total LUTs used by the accelerator.
We evaluate these pareto frontiers with three optimization configurations.
First, we instantiate each layer in each model as a CU without any spatial parallelism optimization and visualize the pareto frontiers in Fig.~\ref{fig:pareto_fronts_lut}a.
Second, we maximize the number of PEs used in each layer in each model and visualize the pareto frontiers in Fig.~\ref{fig:pareto_fronts_lut}b.
Finally, we maximize both the number of PEs and the SIMD lanes used and visualize the pareto frontiers in Fig.~\ref{fig:pareto_fronts_lut}c.

To ensure that the trends we analyze from these estimates are meaningful, we return to the experiments carried out in Section~\ref{sec:motivation}.
We compare the absolute LUT utilization reported from post-synthesis RTL against the corresponding FINN estimates and observe a 94\% correlation.

Reducing the precision of weights and activations provides resource utilization savings in exchange for model performance; however, we observe that adding the accumulator bit width to the design space provides a better overall trade-off.
Our results show that, for a given target accuracy or resource budget, our algorithm can offer a better trade-off between LUT utilization and model performance than existing baselines for various optimization strategies, confirming the benefit of including the accumulator bit width in the overall HW-SW co-design space.

\subsection{Evaluating Resource Savings}
\label{sec:resource_savings}

Because we force the FINN compiler to use LUTs for compute and memory resources wherever possible, we evaluate where our resource savings come from.
To do so, we separate LUT utilization into compute, memory, and control flow.
For compute, we aggregate the LUTs used for adder trees, MACs, and comparison logic; for memory, the LUTs used to store weights, thresholds, and intermediate representation; and for control flow, the LUTs used for on-chip interconnects and AXI-Stream protocols.
In Fig.~\ref{fig:pareto_fronts_breakdown}, we visualize this break down for each of the pareto optimal models that correspond to our pareto frontier in Fig.~\ref{fig:pareto_fronts_lut}.

We observe that the majority of LUT savings come from reductions to memory resources when accelerators are generated without spatial parallelism, but primarily come from compute resources as parallelism is increased.
Without parallelism, the reductions in LUT utilization are largely from the reduced storage costs of thresholds and intermediate activations, which are directly impacted by the precision of the accumulator and output activations.
As parallelism is increased, the reductions in compute LUTs primarily come from the reduced cost of MACs, which are directly impacted by the precision of the weights, inputs, and accumulators.

Finally, we observe that the control flow LUTs largely remain constant for each network, which is expected as the network architecture is not impacted by changes to the data types used.
Noticeably, the relative share of LUTs contributed by control flow logic is higher for networks with skip connections (\textit{e.g.}, ResNet18 and UNet) than without (\textit{e.g.}, MobileNetV1 and ESPCN), and is relatively less impactful as parallelism is increased.

\subsection{A Deeper Look at the Impact of Our Constraints}
\label{sec:sparsity}

\begin{figure}[t!]
\centering
\includegraphics[width=\linewidth]{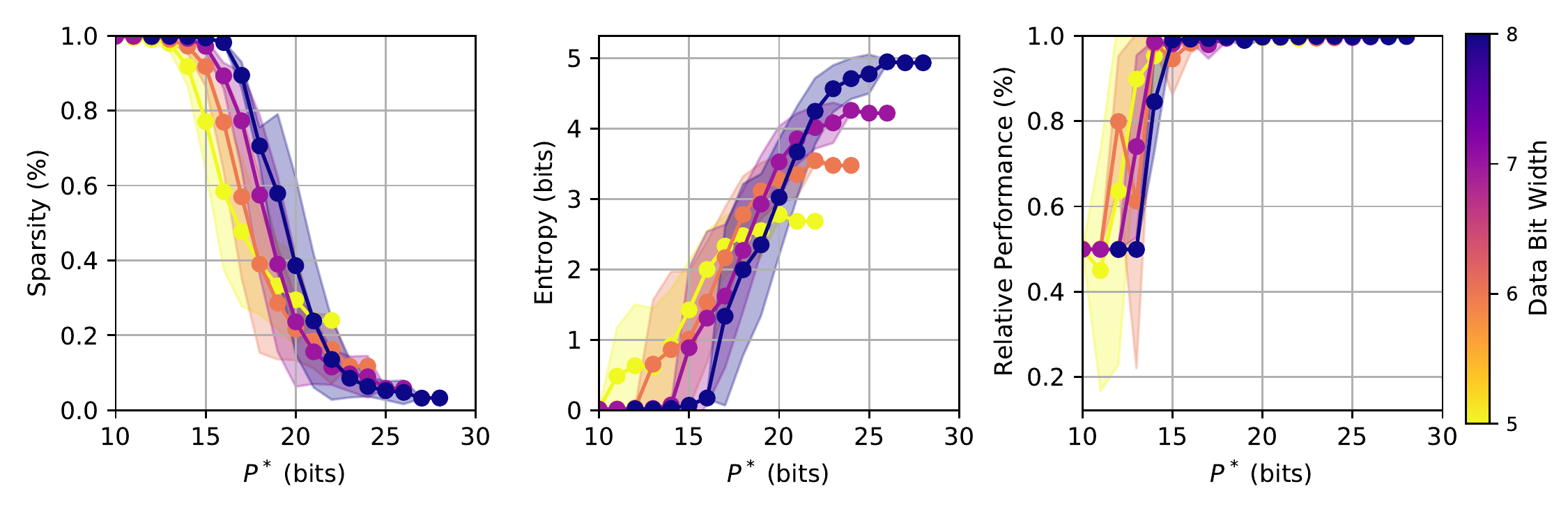}
\caption{As a result of our $\ell_1$-norm constraints, reducing the accumulator bit width exposes opportunities to exploit unstructured sparsity (left) and weight compression (middle) without sacrificing model performance relative to the floating-point baseline (right).}
\label{fig:sparsity}
\end{figure}

As a byproduct of our weight normalization formulation, our quantization algorithm provides a means of not only constraining weights to fit into an accumulator of user-defined bit width, but also of increasing the sparsity and compressibility of the resulting weights, as shown in Fig.~\ref{fig:sparsity}.

\textbf{Sparsity} is the proportion of zero-valued elements in a tensor.
The most common use of sparsity in machine learning workloads is to accelerate inference by reducing compute and memory requirements~\cite{gale2020sparse}.
We direct the interested reader to~\cite{hoefler2021sparsity} for a recent survey of prior work on sparsity in deep learning.
Among this work are various studies regarding the use of $\ell_1$-norm weight regularization as a means of introducing sparsity~\cite{yang2019structured,chao2020directional}.
Our quantization algorithm has a similar effect.
By replacing the $\ell_2$-norm of the standard weight normalization formulation with our log-scale $\ell_1$-norm parameter, we introduce a novel means of encouraging unstructured weight sparsity.
Recall that the value of $P^*$ is used by Eq.~\ref{eq:weight_bound_reparameterized} to upper bound the $\ell_1$-norm of the weights.
Consequently, reducing $P^*$ further constrains this upper bound to encourage weight sparsity as a form of $\ell_1$-norm weight regularization.
In Fig.~\ref{fig:sparsity} on the left, we visualize the average sparsity across all of our benchmark models as we reduce the accumulator bit width $P^*$.

\textbf{Compressibility} is often estimated using the theoretical lower bound on the amount of bits per element as measured by the entropy~\cite{shannon1948mathematical}.
Reducing the entropy reduces the amount of information required for lossless compression, increasing the compression rate.
Prior work has studied the use of entropy regularization as a means of improving weight compression~\cite{agustsson2017soft,aytekin2019compressibility}.
We observe that our $\ell_1$-norm constraints have a similar effect as these techniques.
As shown in Fig.~\ref{fig:sparsity} in the middle, we observe that the entropy decreases as we reduce the accumulator bit width $P^*$.

In Fig.~\ref{fig:sparsity}, we visualize how the sparsity, compressibility, and relative model performance are effected by reductions to $P^*$ using the models from our grid search described in Section~\ref{sec:exp_details}.
To simplify our analysis, we focus on configurations where the weight and input activation bit widths were the same (\textit{e.g.}, $M = N$), and plot the averages observed across all 4 of our benchmark models.
For models with 8-bit weights and activations, we observe that reducing $P^*$ to 16 bits yields an average sparsity of 98.2\% with an estimated compression rate of 46.5x while maintaining 99.2\% of the floating-point performance.

\section{Conclusion \& Future Work}
\label{sec:conclusion}

We propose a novel quantization algorithm to train QNNs for low-precision accumulation.
Our algorithm leverages weight normalization as a means of constraining learned parameters to fit into an accumulator of a pre-defined bit width.
Unlike previous work, which has sought to merely reduce the risk of overflow or mitigate its impact on model accuracy, our approach guarantees overflow avoidance.

Our study is the first to our knowledge that explores the use of low-precision accumulators as a means of improving the design efficiency of programmable hardware used as QNN inference accelerators.
As such, we theoretically evaluate overflow and derive comprehensive bounds on accumulator bit width with finer granularity than existing literature.
Our experiments show that using our algorithm to train QNNs for AMD-Xilinx FPGAs improves the trade-offs between resource utilization and model accuracy when compared to the standard baseline.

Our results inform the following takeaways:
\begin{itemize}[noitemsep,nolistsep]
\item While reducing the size of the accumulator invariably degrades model accuracy, our algorithm significantly alleviates this trade-off.
\item Using our algorithm to train QNNs for lower precision accumulators yields higher performing models for the same resource budget when compared to the baseline.
\item Without spatial parallelism, the majority of our resource savings come from reductions to memory requirements because reducing the accumulator bit width also reduces the cost of storing thresholds and intermediate activations.
\item As spatial parallelism is increased, reductions in compute costs dominate our resource savings because reducing the accumulator bit width reduces the cost of creating more MACs.
\item Our algorithm inherently encourages extreme unstructured sparsity and increased compressibility of the resulting weights of the QNN while maintaining performance relative to the floating-point baseline.
\end{itemize}

The flexibility of FPGAs is a double-edged sword.
The bit-level control allows for the precisions of weights, activations, and now accumulators to be individually tuned for each layer in a QNN; however, the design space exposed by so many degrees of freedom introduces a complex optimization problem.
Our algorithm increases the flexibility of HW-SW co-design by exposing the accumulator bit width as yet another parameter that can be tuned when simultaneously optimizing QNNs and their corresponding inference accelerators.
In future work, we hope to explore the use of state-of-the-art neural architecture search algorithms as a means of navigating this large design space more efficiently.

\section*{Acknowledgements}
We would like to thank Gabor Sines, Michaela Blott, Nicholas Fraser, Yaman Umuroglu, Thomas Preusser, Mehdi Saeedi, Alex Cann, and the rest of the AMD Software Technology, Architecture, and AECG Research teams for insightful discussions and infrastructure support.

\noindent © 2023 Advanced Micro Devices, Inc.  All rights reserved.
AMD, the AMD Arrow logo, Radeon, and combinations thereof are trademarks of Advanced Micro Devices, Inc.
Other product names used in this publication are for identification purposes only and may be trademarks of their respective companies.

\bibliography{references}
\bibliographystyle{mlsys2022}

\end{document}